\documentclass[lettersize,journal]{IEEEtran}
\usepackage{amsmath,amsfonts}
\usepackage{algorithmic}
\usepackage{algorithm}
\usepackage{array}
\usepackage[caption=false]{subfig}
\usepackage{textcomp}
\usepackage{stfloats}
\usepackage{url}
\usepackage{verbatim}
\usepackage{graphicx}

\usepackage[utf8]{inputenc} 
\usepackage[T1]{fontenc}    
\usepackage{url}            
\usepackage{booktabs}       
\usepackage{amsfonts}       
\usepackage{nicefrac}       
\usepackage{microtype}      
\usepackage{xcolor}         

\usepackage{makecell}
\usepackage{amsmath}
\usepackage{accents}
\usepackage{statex}
\usepackage[normalem]{ulem}
\usepackage{enumitem}
\usepackage{multirow}

\usepackage{changepage}
\usepackage{bm}
\usepackage{mathrsfs}

\usepackage{diagbox}
\usepackage{pdftexcmds}
\usepackage{catchfile}
\usepackage{ifluatex}
\usepackage{ifplatform}
\usepackage{threeparttable}

\usepackage{comment}
\usepackage{aecompl}
\usepackage{lipsum}   
\begin{document}

\title{DiffMove: Group Mobility Tendency Enhanced Trajectory Recovery via Diffusion Model}

\author{Qingyue Long, Can Rong, Huandong Wang,~\IEEEmembership{Member,~IEEE,} Shaw Rajib, Yong Li,~\IEEEmembership{Senior Member,~IEEE}
\thanks{Manuscript received xxxx, 2025; revised xxxx, 2025.}
\thanks{Qingyue Long, Can Rong, Huandong Wang, and Yong Li are with the Department of Electronic Engineering, Beijing National Research Center for Information
Science and Technology (BNRist), Tsinghua University, Beijing 100084, China (e-mail: longqy21@mails.tsinghua.edu.cn; rc20@mails.tsinghua.edu.cn;
wanghuandong@tsinghua.edu.cn;
liyong07@tsinghua.edu.cn)}
\thanks{Rajib Shaw is with the Graduate School of Media and Governance, Keio University, Kanagawa 252-0882, Japan (e-mail: shaw@sfc.keio.ac.jp)}
}

\markboth{IEEE TRANSACTIONS ON MOBILE COMPUTING,~Vol.~14, No.~8, March~2025}%
{Shell \MakeLowercase{\textit{et al.}}: A Sample Article Using IEEEtran.cls for IEEE Journals}

\IEEEpubid{0000--0000~\copyright~2025 IEEE}

\maketitle

\begin{abstract}
In the real world, trajectory data is often sparse and incomplete due to low collection frequencies or limited device coverage. Trajectory recovery aims to recover these missing trajectory points, making the trajectories denser and more complete. However, this task faces two key challenges: 1) The excessive sparsity of individual trajectories makes it difficult to effectively leverage historical information for recovery; 2) Sparse trajectories make it harder to capture complex individual mobility preferences. To address these challenges, we propose a novel method called \textbf{DiffMove}. 
Firstly, we harness crowd wisdom for trajectory recovery. Specifically, we construct a group tendency graph using the collective trajectories of all users and then integrate the group mobility trends into the location representations via graph embedding. This solves the challenge of sparse trajectories being unable to rely on individual historical trajectories for recovery. Secondly, we capture individual mobility preferences from both historical and current perspectives. Finally, we integrate group mobility tendencies and individual preferences into the spatiotemporal distribution of the trajectory to recover high-quality trajectories. 
Extensive experiments on two real-world datasets demonstrate that DiffMove outperforms existing state-of-the-art methods. Further analysis validates the robustness of our method.
\end{abstract}


\section{introduction}

Human mobility data reveals people's activity patterns and mobility intentions, making it of significant value in various fields such as epidemic control~\cite{balcan2011phase,yang2022epimob,zhang2022epidemic}, urban planning~\cite{gao2015spatio,kung2014exploring,bengtsson2015using,li2022disenhcn}, and transportation engineering~\cite{liu2019identifying,yue2014zooming}.
However, the sparsity of real-life trajectory data, whether due to low sampling rates of mobile data devices or privacy constraints imposed by users, presents significant challenges and negatively impacts the performance of downstream applications. This limitation has a significant impact in fields such as urban planning and traffic management, as accurate estimates of hourly crowd flows are crucial for effective scheduling and responsive urban management~\cite{li2013efficient}. As a result, trajectory recovery is essential for improving the accuracy and utility of applications in these domains.

Formally, the task of trajectory recovery involves restoring missing trajectory records based on existing partial trajectory data. Throughout the evolution of the technology, traditional methods have primarily relied on rule-based frameworks~\cite{hoteit2013estimating,hoteit2014estimating,gonzalez2008understanding,isaacman2012human,song2010limits}, with the fundamental assumption that human mobility patterns can be described by physical rules (such as movement speed constraints or geographically reachable areas). These methods focus on modeling basic features, such as analyzing the correlation between neighboring trajectory points using the principle of spatial continuity~\cite{xiao2014lightweight,ficek2012inter}, or predicting recurring trajectory patterns based on time periodicity assumptions~\cite{yu2018using}. While these approaches may be feasible in simple scenarios, their simplified modeling often fails to fully capture the diversity and complexity of real-world human mobility, particularly when capturing personalized behaviors, which can result in significant bias.
With technological advances, data-driven methods have gradually become mainstream in research. These methods break free from the limitations of traditional rules and use tools such as deep learning models~\cite{xi2019modelling}, attention mechanisms~\cite{xia2021attnmove}, and variational autoencoders~\cite{long2023VAE} to automatically extract multidimensional features from vast amounts of movement data, including implicit spatiotemporal correlations, behavioral habits, and higher-order semantic patterns. This end-to-end learning paradigm significantly enhances the model's ability to adapt to complex scenarios, thereby improving the accuracy and robustness of trajectory recovery.
Nevertheless, these data-driven methods focus on regular mobility patterns and often neglect the specific mobility preferences embedded in trajectory data, which can significantly affect the quality of trajectory recovery.

Therefore, developing a trajectory recovery method that simultaneously integrates both group commonalities and individual characteristics has become a key solution to enhance the practical value of trajectory recovery. However, constructing such a method faces the following critical challenges:
\IEEEpubidadjcol
\begin{itemize}[leftmargin=*]
\item \textbf{Excessive sparsity in individual trajectories severely limits the effective use of historical information for recovery.} When trajectories are overly sparse, spatiotemporal dependencies can only be inferred from a limited number of data points, which impedes the accurate identification of past mobility patterns. As a result, this limitation significantly complicates the precise recovery of trajectories.

\item \textbf{Sparse trajectories make it harder to capture the dynamic and varied mobility preferences of individuals.} People’s preferences for different locations can differ greatly, often reflecting their attachment to specific places. However, individuals typically weigh multiple factors when choosing travel destinations. For example, a person planning to shop may prioritize malls and compare their preferences among them. This interplay of preferences and behaviors poses a significant challenge for accurate trajectory modeling.

\end{itemize}

To address the above challenges, we propose a method named \textbf{DiffMove}, which consists of three critical modules, \textit{i.e.}, 1) group tendency extraction module, 2) individual preference extraction module, and 3) mobility behavior diffusion module.
In the group tendency extraction module, we construct a group tendency graph using all trajectories. This graph uses the transition frequency between locations as the edge weight, representing the degree of group tendency towards each location. Then, we integrate this group tendency information into the location embeddings through graph embedding techniques, considering both first-order and second-order proximities. This approach effectively addresses the first challenge.
Given the potential noise and sparsity in trajectories, the individual preference extraction module improves accuracy by incorporating historical trajectories related to the current trajectory to capture individual mobility preferences better. We first extract historical preferences from past trajectories using a history processor and current preferences from current trajectories with a current processor. These preferences are then combined using the proposed inter-trajectory attention mechanism.
In the mobility behavior diffusion module, we use a carefully designed diffusion model to integrate individual mobility preferences into the distribution modeling of mobility behaviors. This step ensures the precise capture of individual mobility preferences, thus addressing the second challenge.
Given that group mobility tendency and individual mobility preferences influence mobility behavior, we aim to recover high-quality trajectories by integrating both factors.

Overall, our contributions can be summarized as follows:
\begin{itemize}[leftmargin=*]
\item We propose a group mobility tendency enhanced trajectory recovery method named DiffMove, which leverages a diffusion model to simultaneously capture group tendencies and individual preferences.
\item To capture group mobility tendencies, we construct a group tendency graph and integrate this information into location representations using graph embedding techniques. For modeling individual mobility preferences, we extract preferences from both historical and current trajectories. Finally, leveraging a carefully designed diffusion model, we effectively integrate group tendencies and individual preferences, thereby modeling the distribution of mobility behavior.
\item Extensive experiments on two real-world datasets demonstrate the effectiveness of our DiffMove model. Further analysis confirms the robustness of our approach.
\end{itemize}
\section{related work}
\subsection{Trajectory Recovery}
Trajectory recovery research has evolved along two principal methodological axes: rule-based and data-driven approaches. Rule-based methods are based on the idea that people move around in space and time in clear ways that can be formalized through models that can be understood physically. A prominent implementation involves map-matching techniques that constrain mobility trajectories within road network topologies. For instance, Xiao et al. developed a conditional random field framework incorporating contextual relationships between cellular signaling records (CDR) to reconstruct urban mobility paths~\cite{xiao2014lightweight}. Complementary strategies employ interpolation mechanisms, where missing trajectory points are estimated through linear or Gaussian process regression, capitalizing on spatiotemporal correlations in mobility data~\cite{hoteit2014estimating}~\cite{yu2018using}. However, these methods have a fundamental flaw: they don't consider the nonlinear dynamics in real-world mobility behaviors because they rely on oversimplified motion heuristics.

Data-driven methodologies employ machine learning frameworks to autonomously extract complex mobility patterns from empirical movement datasets, capturing nonlinear temporal dependencies and high-order behavioral correlations. Contemporary implementations include deep learning architectures that model sequential location transitions through recurrent neural networks with attention mechanisms. For instance, Feng et al. developed a hybrid architecture integrating long short-term memory (LSTM) modules with periodic pattern recognition to reconcile historical trajectory regularities with real-time movement predictions~\cite{feng2018deepmove}. The recently proposed Bi-STDDP model advances this paradigm by establishing bidirectional spatiotemporal attention mechanisms that jointly optimize user-specific behavioral vectors and geographic contextual features~\cite{xi2019modelling}. Nevertheless, two fundamental limitations persist: (1) the absence of prior knowledge about human mobility patterns leads to inefficient feature extraction from noisy observational data; (2) overreliance on historical trajectory patterns inadequately addresses sparse data scenarios, failing to recover latent mobility semantics.

In general, rule-based and data-driven methods encounter difficulties in effectively resolving the challenges associated with trajectory recovery. To mitigate the problem of trajectory sparsity, we introduce a novel recovery strategy leveraging a diffusion model. This method can effectively use historical information, population information, modeling mobility periodicity, mobility regularity, and spatial continuity.

\subsection{Diffusion Models}
Most studies on diffusion models build upon Denoising Diffusion Probabilistic Models (DDPM), which employ variational inference. The fundamental idea of DDPM is to gradually modify the data distribution through a stepwise diffusion mechanism and subsequently reconstruct it by learning an inverse transformation. This method enables the development of a versatile and computationally efficient generative framework~\cite{ho2020denoising}. The diffusion model has shown impressive performance in image synthesis tasks~\cite{wijmans1995solution,ho2022video,gu2022vector,ruiz2023dreambooth,kawar2023imagic,rombach2022high}.


Due to its strong generative abilities, the diffusion model has expanded beyond image synthesis to audio synthesis. The DiffWave model is an efficient vocoder that delivers high-quality speech with a small footprint and is faster than real-time generation. It has also led to significant advancements in category-based labeling and unconditional audio generation~\cite{kong2020diffwave}.

The diffusion model has also found applications in time series modeling. For time series imputation, CSDI~\cite{tashiro2021csdi} introduced a novel interpolation method based on a fraction-based diffusion model. This approach leverages self-supervised training to optimize the model, allowing it to capture temporal correlations effectively. Similarly, SSSD~\cite{alcaraz2022diffusion} employs a conditional diffusion model in conjunction with a structured state space framework to address extended temporal dependencies in time series, demonstrating strong capabilities in both interpolation and prediction. In the context of time series forecasting, TimeGrad~\cite{rasul2021autoregressive} proposes an autoregressive approach for generating multivariate probabilistic time series by sequential sampling from the underlying data distribution. This approach leverages a diffusion-based probabilistic model, drawing upon principles related to energy-based methods.

Overall, diffusion models have generally performed well in time series modeling, audio synthesis, and picture synthesis. However, limited research explores the application of diffusion models to trajectory recovery. This task differs from general time series modeling as it requires the consideration of mobility-specific features inherent to trajectories.

\section{preliminaries}
This section presents the concept and process of the denoising diffusion probabilistic model after introducing the concepts and notations used in this work.

\begin{table}[t]
\caption {{Summary of Notations.}}
\centering 
\setlength{\tabcolsep}{2mm}
{
\begin{tabular}{c|p{6cm}}
\toprule  
Notation& Description\\
\midrule  
$u$& A user.\\
\hline
\multirow{2}*{\makecell{$l_{u}^{i,n}$}}& The location of user $u$ within a specified time period on the $n$-th time slot of the $i$-th day.\\
\hline
\multirow{2}*{\makecell{$\mathcal{T}_{u}^{i}$, $\mathcal{T}_{u}^{I}$}}& The historical trajectory of user $u$ on day $i$ and the current trajectory of user $u$.\\
\hline
$G$&  A group preference graph.\\
\hline
$e_l$&  The embedding of location $l$.\\
\hline
$e_t$&  The temporal representation of the $n$-th time interval.\\
\hline
\multirow{2}*{\makecell{$\bar{e}_{u}^{i, n}$}}&  The temporal-aware representation corresponding to the $n$-th time interval within the $i$ trajectory of user $u$.\\
\hline
\multirow{3}*{\makecell{$\alpha_{n, k}^{(h)}$}}&  The degree of similarity computed by the $h$-th head between time interval  $n$ in the current trajectory and time interval $k$ in the historical trajectory.\\
\hline
\multirow{2}*{\makecell{$\check{e}_{u}^{p, n}$}}&  The historical shift-aware representation of the $n$-th time interval in the $p$-th trajectory of user $u$.\\
\hline
\multirow{2}*{\makecell{$\check{e}_{u}^{i, n}$}}&  The current shift-aware representation of the $n$-th time interval in the $i$-th trajectory of user $u$.\\
\hline
\multirow{2}*{\makecell{${e}_{u}^{n}$}}&  The final trajectory representation of the $n$-th time interval.\\
\hline
\multirow{2}*{\makecell{${{e}_{u}^{n}}_0$}}&  The clean final trajectory representation of the $n$-th time interval.\\
\hline
${{e}_{u}^{n}}_t$&  The representation in a noising step $t$.\\
\hline
${\hat{e}_{u}^{n}}_0$&  The clean sample.\\
\hline
$\mathcal{F}_\theta(\cdot)$&  The denoising network.\\
\hline
$\mathcal{L}_{simple}$, $\mathcal{L}_d$&  The simplified loss and distance-aware loss.\\
\hline
$\lambda_d$&  The weight of distance loss.\\
\bottomrule 
\end{tabular}
}
\label{table:notation}
\end{table}

\subsection{Problem definition}
\textit{Definition 1 (Trajectory).} A trajectory represents the sequence of locations visited by a user over time. The trajectory of user $u$ on day $i$ can be expressed as a set of spatiotemporal coordinates, denoted by $\mathcal{T}_{u}^{i} = {l{u}^{i, 1}, ..., l_{u}^{i, n}, ..., l_{u}^{i, N}}$, where $l_{u}^{i, n}$ indicates the user's location at the $n$-th time slot within a specified interval (e.g., every 30 minutes), and $N$ is the total number of time slots. It is important to note that if the location for a particular time slot $n$ is not observed, the value $l_{u}^{i, n}$ is marked as \textit{null}, signifying a missing location.
                                        
\textit{Definition 2 (Current and Historical Trajectory).}
Let $I$ represent a target day, and $\mathcal{T}_{u}^{I}$ denote the trajectory of user $u$ on that day. This trajectory $\mathcal{T}_{u}^{I}$, corresponds to the user's current path, while the historical trajectories are denoted as $\{\mathcal{T}_{u}^{1}, \mathcal{T}_{u}^{2}, \dots, \mathcal{T}_{u}^{I-1}\}$, representing the user's movement on the days before day $I$.

\textit{Problem Statement (Trajectory Recovery).}
The goal is to reconstruct the full trajectory for the current day by filling in the missing locations or null values in $\mathcal{T}_{u}^{I}$, leveraging the historical trajectories $\{\mathcal{T}_{u}^{1}, \mathcal{T}_{u}^{2}, \dots, \mathcal{T}_{u}^{I-1}\}$ associated with user $u$.

\subsection{Denoising Diffusion Probabilistic Model}
Diffusion models can be described within the latent variable framework, characterized as $p_\theta(x_0) := \int p_\theta(x_{0:T}) , dx_{1:T}$, in which the latent variables $x_1, \dots, x_T$ have dimensions identical to the observed data $x_0 \sim q(x_0)$. This probabilistic formulation utilizes two distinct Markov processes: a forward diffusion process progressively injecting noise into the original input, and a reverse denoising process reconstructing the original input by systematically removing noise. The detailed diffusion procedure guided by these Markov processes is described below:
\begin{equation}\label{equ:DDPM1}
q\left(\mathbf{x}_{1: T} \mid \mathbf{x}_{0}\right):=\prod_{t=1}^{T} q\left(\mathbf{x}_{t} \mid \mathbf{x}_{t-1}\right) 
\end{equation}
where $q\left(\mathbf{x}_{t} \mid \mathbf{x}_{t-1}\right):=\mathcal{N}\left(\sqrt{1-\beta_{t}} \mathbf{x}_{t-1}, \beta_{t} \mathbf{I}\right)$ and $\beta_t$ is a minor positive constant indicative of the noise intensity.  Similarly, $x_t$ may be articulated in closed form as $x_{t}=\sqrt{\alpha_{t}} x_{0}+\left(1-\alpha_{t}\right) \epsilon$ for $\epsilon \sim \mathcal{N}(0, \mathbf{I})$, where $\alpha_{t}=\sum_{i=1}^{t}\left(1-\beta_{t}\right)$.

In contrast, the goal of the reverse procedure is to progressively remove noise from $x_t$, thereby reconstructing the original sample $x_0$. This denoising operation can be represented through the following Markov chain:
\begin{equation}\label{equ:DDPM2}
p_{\theta}\left(\mathbf{x}_{0: T}\right):=p\left(\mathbf{x}_{T}\right) \prod_{t=1}^{T} p_{\theta}\left(\mathbf{x}_{t-1} \mid \mathbf{x}_{t}\right)
\end{equation}
where $\mathbf{x}_{T} \sim \mathcal{N}(\mathbf{0}, \mathbf{I})$. Additionally, it is assumed that $p\theta(x_{t-1} \mid x_t)$ follows a normal distribution, characterized by learnable parameters as follows:
\begin{equation}\label{equ:DDPM3}
p_{\theta}\left(\mathbf{x}_{t-1} \mid \mathbf{x}_{t}\right):=\mathcal{N}\left(\mathbf{x}_{t-1} ; \boldsymbol{\mu}_{\theta}\left(\mathbf{x}_{t}, t\right), \sigma_{\theta}\left(\mathbf{x}_{t}, t\right) \mathbf{I}\right)
\end{equation}
Ho et al.~\cite{ho2020denoising} introduced denoising diffusion probabilistic models (DDPM), which employ a distinct parameterization for $p_\theta(x_{t-1} \mid x_t)$ as follows:
\begin{equation}\label{equ:DDPM4}
\begin{cases}
\boldsymbol{\mu}_{\theta}\left(\mathbf{x}_{t}, t\right) = \frac{1}{\alpha_{t}}\left(\mathbf{x}_{t}-\frac{\beta_{t}}{\sqrt{1-\alpha_{t}}} \boldsymbol{\epsilon}_{\theta}\left(\mathbf{x}_{t}, t\right)\right), \\
\sigma_{\theta}\left(\mathbf{x}_{t}, t\right) = \tilde{\beta}_{t}^{1 / 2} \text{ where } \tilde{\beta}_{t} = 
\begin{cases}
\frac{1-\alpha_{t-1}}{1-\alpha_{t}} \beta_{t} & t > 1, \\
\beta_{1} & t = 1.
\end{cases}
\end{cases}
\end{equation}
where $\epsilon_\theta$ denotes a trainable denoising function. The reverse procedure can be optimized by minimizing the following training objective:
\begin{equation}\label{equ:DDPM6}
\min _{\theta} \mathcal{L}(\theta):=\min _{\theta} \mathbb{E}_{\mathbf{x}_{0} \sim q\left(\mathbf{x}_{0}\right), \boldsymbol{\epsilon} \sim \mathcal{N}(\mathbf{0}, \mathbf{I}), t}||\boldsymbol{\epsilon}-\boldsymbol{\epsilon}_{\theta}\left(\mathbf{x}_{t}, t\right)||_{2}^{2},
\end{equation}
where $\mathbf{x}_{t}=\sqrt{\alpha_{t}} \mathbf{x}_{0}+\left(1-\alpha_{t}\right) \boldsymbol{\epsilon}$. The denoising function $\epsilon_\theta$ produces results by estimating the noise vector applied to the stochastic input $x_t$. This objective can be interpreted as a weighted variational bound on the negative log-likelihood, assigning less importance to terms corresponding to smaller values of $t$, especially those involving minimal noise levels.
\section{method}
\begin{figure*}[t]
\centering
\includegraphics[width=1.0\textwidth]{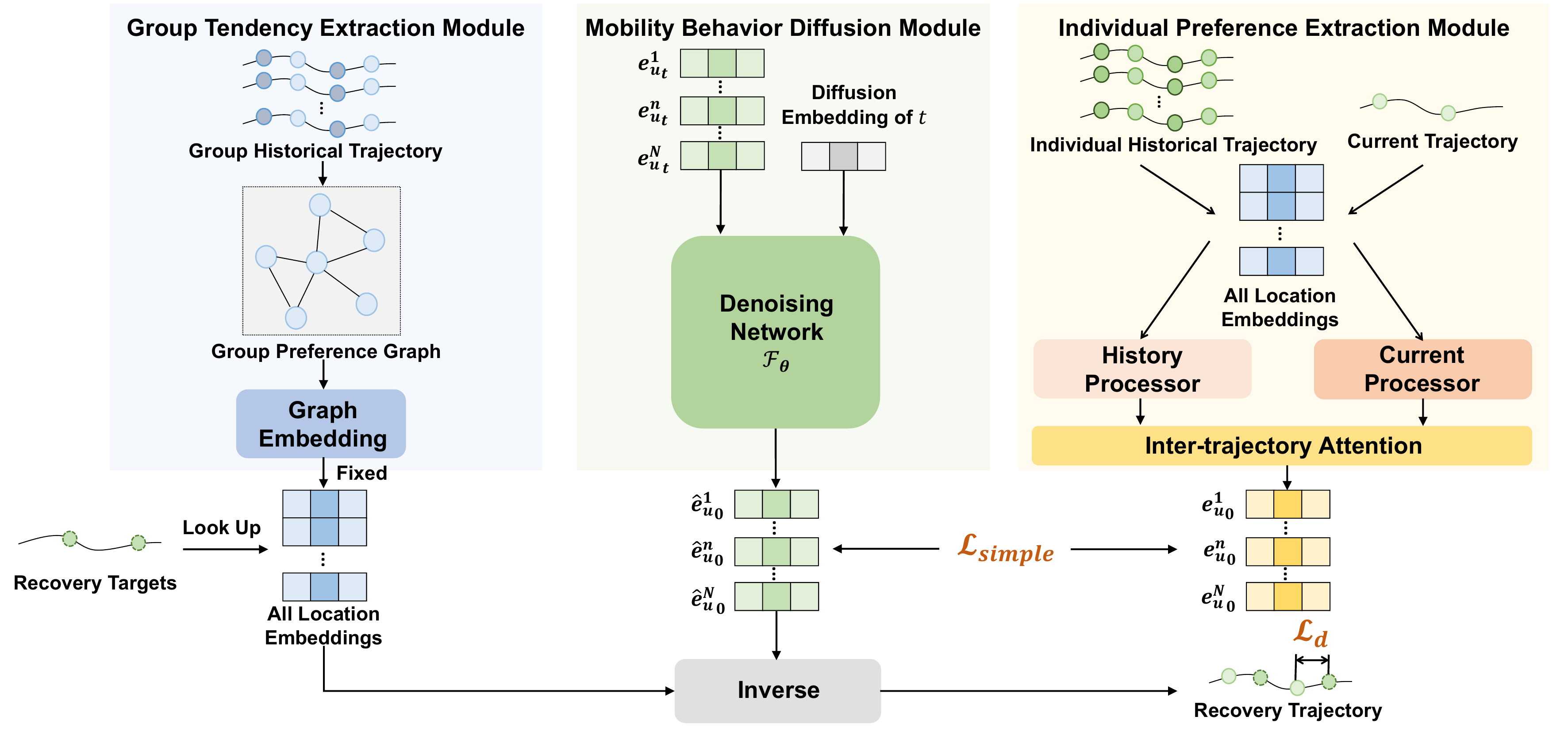}
\caption{The framework of DiffMove.}
\label{fig:framework}
\end{figure*}

The framework illustrated in Figure~\ref{fig:framework} depicts our proposed DiffMove model, which consists of three main components: 1) a group tendency extraction module, 2) an individual preference extraction module, and 3) a mobility behavior diffusion module. We provide detailed descriptions of these components in this section.

\subsection{Group Tendency Extraction Module}\label{sec::Group Preference Extraction Module}
To address challenge 1, we propose to construct a \textbf{group tendency graph} and employ appropriate graph embedding techniques to accurately capture the collective tendencies of the population regarding location transitions (i.e., travel origins and destinations).

\subsubsection{Graph Construction}
The frequency of travel to a specific location reflects the group's tendency to visit that destination. For example, when a location appears frequently across multiple trajectories, it indicates a shared preference for that destination. To effectively model this collective behavior, we propose the construction of a group tendency graph, denoted as $G$, using all available trajectory data. Formally, $G$ is defined as $G = (V, E)$, where $V$ represents the set of nodes corresponding to all locations visited by the group, and $E$ denotes the set of edges that represent the relationships between pairs of locations. Each edge $e \in E$ is an unordered pair $e = (u, v)$ with an associated weight $w_{uv} > 0$, where $w_{uv}$ captures the frequency of transitions between locations $u$ and $v$. This weight reflects the group's stronger inclination to travel between $u$ and $v$. Through this approach, the group tendency graph $G$ accurately represents the mobility preferences of the population across various locations.

\subsubsection{Group Tendency Extraction with Graph Embedding}
Since the edge weight $w_{uv}$ in the group tendency graph reflects the number of transitions between locations $u$ and $v$, which represents the population's inclination towards these destinations, we aim to incorporate this group tendency data into the location embeddings through a graph embedding module. However, the edges representing group tendencies may suffer from sparsity in the constructed graph. This sparsity arises due to the common herd effect, where large crowds primarily move within a few popular areas. As a result, the transition frequencies—and consequently the edge weights—are generally higher among these popular locations, leading to challenges in graph learning. To counteract the impact of this herd effect on modeling group tendencies and to better account for less frequented locations, we propose incorporating group tendency information from two perspectives: \textbf{first-order proximity} and \textbf{second-order proximity}.

To effectively perform embedding, it is essential to preserve the structure of the graph network. First, maintaining the local network structure, specifically the pairwise proximity between nodes, is critical. The first-order proximity between nodes can define the local graph structure:

\textbf{1) First-order Proximity.} In the group tendency graph $G$, for every edge $(u, v)$ between two nodes, the associated weight $w_{uv}$ represents the first-order proximity between $u$ and $v$. This weight quantifies the strength of the group's tendency to travel between these two locations, effectively capturing local relationships. However, due to the herd effect in group tendency modeling, where popular locations dominate, the graph suffers from sparsity. As a result, relying only on first-order proximity does not fully preserve the graph's structure. A natural extension is to consider nodes with common neighbors, as these nodes are likely to exhibit similar characteristics. To address the sparsity problem, we introduce second-order proximity.

\textbf{2) Second-order Proximity.}
Second-order proximity between nodes $(u, v)$ captures how closely their neighboring structures resemble each other. Formally, for a node $u$, we define its first-order relationships with all nodes as a vector $p_u = (w_{u,1}, \dots, w_{u,|V|})$. Then, the second-order proximity between nodes $u$ and $v$ is quantified by assessing the similarity of their respective vectors $p_u$ and $p_v$. Incorporating second-order proximity highlights less frequently visited locations (nodes with lower edge weights) and addresses the sparsity challenge by modeling indirect neighborhood information.

\textbf{3) Group Tendency Extraction.} 
Next, we aim to integrate group tendency information into location embeddings using graph embedding techniques. Given the constructed group tendency graph $G = (V, E)$, the goal of graph embedding is to represent each vertex $v \in V$ as a vector in a lower-dimensional space $\mathbb{R}^d$, where $d \ll |V|$. Specifically, we adopt the LINE (Large-scale Information Network Embedding) method~\cite{tang2015line}, a widely used approach for graph embedding in large-scale information networks, to incorporate group tendency information from both first-order and second-order proximity perspectives.
For first-order proximity, we minimize the discrepancy between the predicted and actual edge weights to capture the group tendency information. To address second-order proximity, we perform random walks and negative sampling within the second-order proximity neighborhoods, helping to mitigate issues of sparsity and the herd effect in trajectory data. We separately optimize these two objectives, representing first-order and second-order proximity, using the LINE method. Finally, we combine the embeddings derived from both perspectives to obtain a consolidated embedding for each location.
Using the group tendency graph $G$, the graph embedding process with LINE can be expressed as follows:
\begin{equation}\label{equ:LINE}
e_{l}={\rm LINE}_\theta(G),
\end{equation}
For each location $l \in L$, we assign a representation vector $e_l \in \mathbb{R}^d$. These embeddings collectively form the embedding matrix $E_l \in \mathbb{R}^{|L+1| \times d}$.

In conclusion, we build a group tendency graph based on the available trajectory data. We then utilize the LINE method for graph embedding, incorporating group tendency information into location embeddings by considering both first-order and second-order proximity relationships.

\begin{figure}[t]
\centering
\includegraphics[width=0.48\textwidth]{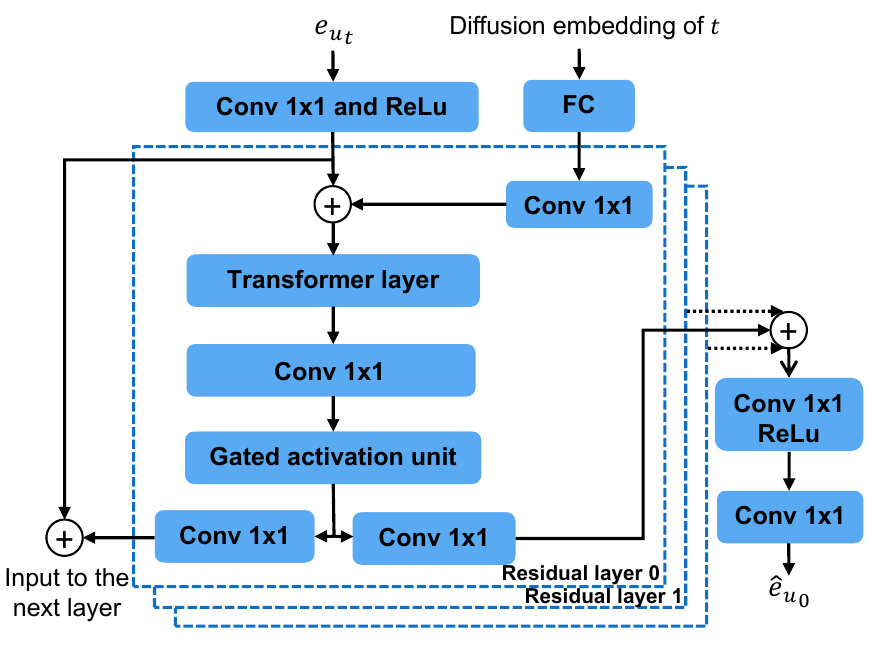}
\caption{The architecture of denoising network $\mathcal{F}_{\theta}$.}
\label{fig:csdi}
\end{figure} 
\subsection{Individual Preference Extraction Module}
To address challenge 2, we propose \textbf{trajectory processor} and \textbf{inter-trajectory attention} to accurately capture individuals' preferences for visiting locations or mobility patterns.

\subsubsection{Trajectory Processor}
The trajectory processor consists of two main parts: the history processor and the current processor. The history processor analyzes multiple past trajectories to derive periodic mobility preferences for each individual. On the other hand, the current processor focuses on identifying the individual's present preferences for location selection, emphasizing the types of places they tend to visit most frequently.

\textbf{History Processor.}
The mobility patterns of each user are complex and variable, making it vital to capture their individual preferences accurately from their trajectories. However, a single trajectory is typically too sparse to reveal sufficient details about a user's favored destinations and mobility behaviors. As a result, we use the history processor to effectively model individual mobility preferences by leveraging comprehensive, long-term historical trajectory data.

To capture spatio-temporal dependencies, we jointly embed time and location into a dense representation as input. Specifically, for each location $l \in L$, we define an embedding vector $e_l \in \mathbb{R}^d$, obtained by constructing a group tendency graph and applying graph embedding through the group tendency extraction module in Section~\ref{sec::Group Preference Extraction Module}.
Temporal information also plays a key role in mobility modeling. Following~\cite{vaswani2017attention}, we generate an embedding for each time slot $t$ as follows:
\begin{equation}\label{equ:embedd1}
\left\{\begin{array}{l}
e_{t}(2 i)=\sin \left(t / 10000^{2 i / d}\right), \\
e_{t}(2 i+1)=\cos \left(t / 10000^{2 i / d}\right),
\end{array}\right.
\end{equation}
where the index $i$ indicates the specific dimension, the embedding vectors representing time have the same dimensionality $d$ as those representing locations.

Ultimately, to obtain a temporal-aware embedding for the $n$-th time interval of $u$’s $i$-th trajectory, denoted by $\bar{e}_{u}^{i, n} \in \mathbb{R}^{d}$, we combine the associated time and location embedding vectors through element-wise addition:
\begin{equation}\label{equ:embedd2}
\bar{e}_{u}^{i, n}=e_{l}+e_{t},
\end{equation}
Additionally, this allows us to perform the subsequent calculation on a vector that has fewer dimensions than the initial one-hot vector.

Since a single historical trajectory contains limited information to capture trajectory periodicity effectively, we design a history aggregator to integrate multiple sparse trajectories. The structure of this aggregator is as follows:
\begin{equation}\label{equ:aggregator}
\mathcal{T}_{u}^{p}=\mathcal{T}_{u}^{1} \oplus \mathcal{T}_{u}^{2} \cdots \oplus \mathcal{T}_{u}^{i-1},
\end{equation}
where $\oplus$ denotes the extraction of the most frequently visited location during the specified time interval. Thereafter, the aggregated $\mathcal{T}_{u}^{p}$ is incorporated into $\bar{e}_u^p$, with each time interval represented by $\bar{e}_u^{p,n}$ as per Eq.(~\ref{equ:embedd2}).

If certain users have no recorded data for a given time slot, the aggregation mechanism alone cannot fill in the missing locations. To address this, we introduce an attention mechanism that infers missing locations based on observations from other time slots. Specifically, a multi-headed attention network is employed to capture the spatio-temporal relationships between trajectory points. Under the attention head $h$, the relationship between time intervals $n$ and $k$ is computed as:
\begin{equation}\label{equ:attention1}
\alpha_{n, k}^{(h)}=\frac{\exp \left(\phi^{(h)}\left(\bar{e}_{u}^{p, n}, \bar{e}_{u}^{p, k}\right)\right)}{\sum_{g=1}^{N} \exp \left(\phi^{(h)}\left(\bar{e}_{u}^{p, n}, \bar{e}_{u}^{p, k}\right)\right)},
\end{equation}

\begin{equation}\label{equ:attention2}
\phi^{(h)}\left(\bar{e}_{u}^{p, n}, \bar{e}_{u}^{p, k}\right)=\left\langle W_{Q}^{1(h)} \bar{e}_{u}^{p, n}, W_{K}^{1(h)} \bar{e}_{u}^{p, k}\right\rangle,
\end{equation}
where $W_{Q}^{1(h)}, W_{K}^{1(h)} \in \mathbb{R}^{d^{\prime} \times d}$ denote transformation matrices, while $<,>$ represents the inner product function.

To obtain the shift-aware representation for the $n$-th time interval in the user $u$'s $p$-th historical trajectory, the similarity $\alpha_{n, k}^{(h)}$ is used to aggregate information across all time slots within the same trajectory.
\begin{equation}\label{equ:attention3}
\widetilde{e}_{u}^{p, n(h)}=\sum_{k=1}^{T} \alpha_{n, k}^{(h)}\left(W_{V}^{1(h)} \bar{e}_{u}^{p, k}\right),
\end{equation}

\begin{equation}\label{equ:attention4}
\widetilde{e}_{u}^{p, n}=\widetilde{e}_{u}^{p, n(1)} \oplus \widetilde{e}_{u}^{p, n(2)} \oplus \cdots \oplus \widetilde{e}_{u}^{p, n(H)}
\end{equation}
where $H$ signifies the total count of heads, $W_{V}^{1(h)} \in \mathbb{R}^{d^{\prime} \times d}$ specifies the transformation matrix, and $\|$ indicates the concatenation operator.

The following standard residual connection is then added for each time period to achieve the shift-aware embedding of historical trajectories:
\begin{equation}\label{equ:attention5}
\check{e}_{u}^{p, n}=\operatorname{Re} L U\left(\widetilde{e}_{u}^{p, n}+W^{1} \bar{e}_{u}^{p, n}\right),
\end{equation}
where $W^{1} \in \mathbb{R}^{d^{\prime} H \times d}$ serves as the projection matrix to address dimension mismatches, while $ReLU(z) = \max(0, z)$ represents the non-linear activation function.

By leveraging the complete history processor, we obtain the vector $\check{e}_{u}^{p, n}$, which encapsulates the mobility characteristics of each individual historical trajectory.

\textbf{Current Processor.}
Users move between various locations throughout the day, with their visit frequency and duration indicating different preference levels. For example, they may regularly spend extended time at residential or workplace locations, whereas visits to shopping malls or entertainment venues tend to be shorter and less frequent. Utilizing current trajectory data, the current processor is designed to model individual location preferences effectively.

First, the trajectory $\mathcal{T}_u^i$ undergoes dense embedding $\bar{e}_u^{i}$ via the previously introduced mobility embedding technique. Subsequently, an attention mechanism is applied$\bar{e}_u^{i}$, following the same network architecture as outlined in Eq.~(\ref{equ:attention1})–(\ref{equ:attention5}), except that the historical pattern $\bar{e}_u^{p}$ is replaced with the current trajectory representation $\bar{e}_u^{i}$. The corresponding projection matrices, denoted as $W_Q^{2}$, $W_V^{2}$, $W_K^{2}$, and $W^{2}$, facilitate the transformation. By stacking multiple layers, this mechanism effectively captures spatial-temporal dependencies, ultimately refining the trajectory representation$\check{e}_{u}^{i, n}$.

\subsubsection{Inter-trajectory Attention}
After comprehensively capturing an individual's past and present preferences, the inter-trajectory attention mechanism evaluates the degree to which the reconstructed trajectory depends on interpolating the observed location versus integrating candidate values derived from historical movements.

The similarity between the current and historical trajectories, denoted as $\alpha$, is defined based on the correlation between their enhanced representations at corresponding time slots, specifically $\check{e}_{u}^{i, n}$ and $\check{e}_{u}^{p, k}$ for all $n, k = 1,2, \dots, N$. The historical candidates are then aggregated according to $\alpha$, with a residual connection incorporated to retain the original interpolation results. The network structure remains consistent with Eqs.~(\ref{equ:attention1})–(\ref{equ:attention5}), employing the projection matrices $W_Q^{3}$, $W_V^{3}$, $W_K^{3}$, and $W^{3}$. Ultimately, the refined representation for the $n$-th time slot, denoted as ${e}_{u}^{n}$, is obtained by integrating $\check{e}_{u}^{p, k}$ and $\check{e}_{u}^{i, n}$.

\subsection{Mobility Behavior Diffusion Module}
Due to the variability and dynamic nature of users' mobility patterns, their immediate preferences often exhibit uncertainty. Thus, representing current mobility behavior as a distribution provides a more effective modeling approach. The diffusion-based method systematically refines this distribution in an iterative manner, enabling a precise characterization of users' mobility behaviors.

Our objective is to capture the distribution of mobility behavior, which is learned through a diffusion model. Specifically, for each time slot's final representation ${e}_{u}^{n}$, the diffusion process can be expressed using a Markov chain, denoted as (${e_u^n}_0$, ${e_u^n}_1$, \dots, ${e_u^n}_t$, \dots, ${e_u^n}_T$), with $T$ indicating the total diffusion steps. The process involves progressively introducing Gaussian noise into the spatial and temporal components from ${e_u^n}_0$ to ${e_u^n}_T$, eventually transforming them into pure Gaussian noise. The diffusion follows the probabilistic framework given by:
\begin{equation}
q\left({e_u^n}_t \mid {e_u^n}_{t-1}\right):=\mathcal{N}\left({e_u^n}_t ; \sqrt{1-\beta_{t}} {e_u^n}_t, \beta_{t} I\right),
\end{equation}
where $\alpha_{t}=1-\beta_{t}$ and $\bar{\alpha}_{t}=\prod_{i=1}^{t}{\alpha_{t}}$.

Rather than directly reconstructing the point $e_u^n$, we formulate its recovery as a reverse denoising process, iteratively refining ${e_u^n}_T$ back to ${e_u^n}_0$. The complete reverse denoising procedure is expressed as follows:
\begin{equation}
p_{\theta}\left({e_u^n}_{0:T} \right):=p\left({e_u^n}_{T}\right) \prod_{t=1}^{T} p_{\theta}\left({e_u^n}_{t-1} \mid {e_u^n}_{t}\right),
\end{equation}

We use a denoising network structure $\mathcal{F}_{\boldsymbol{\theta}}(\cdot)$ similar to DiffWave~\cite{kong2020diffwave} and CSDI~\cite{tashiro2021csdi} to generate ${\hat{e}_u^n}_0$(Figure~\ref{fig:csdi}). 

At the final stage of the reverse process, an additional transformation is applied to map the continuous embedding ${e_u^n}_0$ back to the discrete location ID $l_i$.

\subsection{Training and Sampling}
\subsubsection{Training}
Following the approach proposed in~\cite{ho2020denoising}, we adopt an alternative formulation for signal prediction instead of directly using $\epsilon_t$, as outlined in~\cite{ramesh2022hierarchical}:
$\hat{e_u^n}_0 = \mathcal{F}_\theta(\hat{e_u^n}_t, t)$ with the simple objective~\cite{ho2020denoising} as follows,
\begin{equation}\label{equ:diffusion}
\mathcal{L}_{\text{simple}} = 
\mathbb{E}_{{e_u^n}_0 \sim q({e_u^n}_0), \, t \sim [1, T]}
\parallel {e_u^n}_0 - \mathcal{F}_\theta(\hat{e_u^n}_t, t) \parallel_{2}^{2}.
\end{equation}

Human mobility exhibits inherent regularities, such as spatial continuity. To enforce this property, we introduce a distance-aware loss function, $\mathcal{L}_d$, which constrains the displacement between consecutive mobility steps. Specifically, it measures the Euclidean distances between adjacent points in a trajectory and promotes spatial continuity by minimizing their cumulative sum, as formulated below:
\begin{equation}\label{equ:location loss}
\mathcal{L}_d = \frac{1}{N-1} \sum_{i=1}^{N-1}\parallel{l_{u}^{i,n+1}-l_{u}^{i,n}}\parallel_{2}^{2}.
\end{equation}

Overall, our training loss is
\begin{equation}\label{equ:location loss}
\mathcal{L}= \mathcal{L}_{simple} + \lambda_d \mathcal{L}_d,
\end{equation}
where $\lambda_d$ is the weighting factor of distance-aware loss of $\mathcal{L}_d$.

\subsubsection{Sampling}
\begin{figure}[t]
\centering
\includegraphics[width=0.45\textwidth]{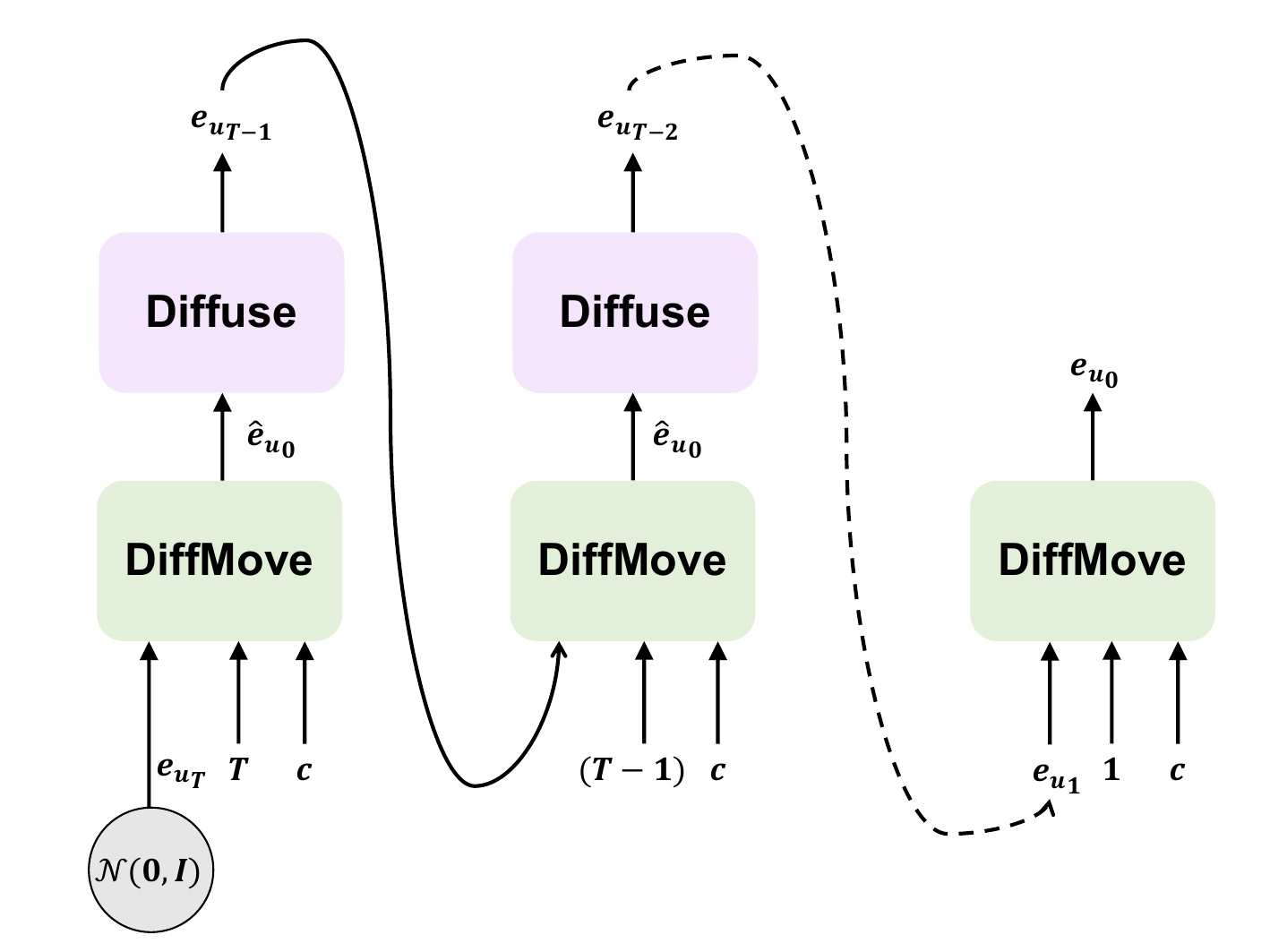}
\caption{The sampling process of DiffMove.}
\label{fig:sample}
\end{figure}

As described in~\cite{ho2020denoising}, the generation of $p({{{e}_u^n}_0})$ follows an iterative process. At each timestep $t$, the model predicts a clean estimate, ${{\hat{e}}_{u}^{n}}_{0} = \mathcal{F}_\theta({{\hat{e}}_u^n}_t, t)$, which is then perturbed to obtain ${{{e}_u^n}_{t-1}}$. This procedure is repeated from $t = T$ down to $t = 0$, progressively refining the representation until ${{{e}_u^n}_0}$ is reconstructed, as illustrated in Figure~\ref{fig:sample}.

\begin{table}[t]
\caption{Basic statistics of mobility datasets.}
\vspace{-10px}
\label{table:datasets}
\begin{center}
\begin{tabular}{ >{\centering\arraybackslash}m{1cm} 
>{\centering\arraybackslash}m{1cm} 
>{\centering\arraybackslash}m{1.2cm} 
>{\centering\arraybackslash}m{1cm}
>{\centering\arraybackslash}m{1cm}
>{\centering\arraybackslash}m{1cm}}
 \hline
 Dataset &City  &Duration &\#Users &\#Loc &\#Traj\\ 
 \hline
 Geolife & Beijing & 5 years & 40 & 3439 & 896\\ 
 Tencent & Beijing & 1 month & 4246 & 8998 & 39422\\ 
\hline
\end{tabular}
\end{center}
\end{table}

\begin{table*}[t]
\center
\caption{Overall performance comparison. The best result in each column is in bold, while the second is underlined.}
\label{tab:performance}
\setlength{\tabcolsep}{1.2mm}{
\scalebox{1.2}{
\begin{tabular}{l|cccccc}
\hline
\multirow{2}{*}{\diagbox[innerwidth=10em]{Algorithm}{Metric}}
    &  
    & \textbf{Geolife}
    & 
    &  
    & \textbf{Tencent}
    & 
    \\
    & Recall   
    & MAP   
    & Distance(m)
    & Recall   
    & MAP      
    & Distance(m)
  \\\hline
Linear	 
    &0.3624 	
    &0.3843	
    &\textbf{2345}	 
    &0.6204 	
    &0.6532	
    &1182
  \\
History	 
    &0.2518 	
    &0.2701 	
    &5107	 
    &0.4702  	
    &0.4896	
    &1687
   \\
Top
    & 0.2732 	
    & 0.2798	
    & 5315	
    & 0.5895  	
    &0.6146	
    &2513
    \\
RF
    &0.2589  	
    &0.2554	
    &6107	
    &0.4832 	
    &0.4904	
    &8138 
    \\
    \hline
LSTM
    & 0.2808  	
    &0.3472	
    &5746	 
    &0.6178  	
    &0.6875	
    &3586
    \\
BiLSTM
    &0.3528  	
    &0.4195	
    &5021	
    &0.7005 	
    &0.7823	
    &1354
    \\
DeepMove
    &0.3403 	
    &0.4123 	
    &4858	 
    &0.7302 	
    &0.7904	
    &1289
    \\
Bi-STDDP
    &0.3765  	
    &0.4514	
    &4032	
    &0.7105	
    &0.7872 	
    &1107
    \\
AttnMove
    &0.3982  	
    &0.4691	
    &3886	
    &0.7646 	
    &0.8249	
    &934
    \\
PeriodicMove
    &\underline{0.4074}  	
    &\underline{0.4805}	
    &3527	
    &\underline{0.7893}  	
    &\underline{0.8501}	
    &\underline{786}
    \\
TrajGDM
    & 0.4016  	
    & 0.4742
    & 3705	
    & 0.7751  	
    & 0.8393	
    & 814
    \\
    \hline
DiffMove
    &\textbf{0.4272}	
    &\textbf{0.5017}	
    &\underline{3126}	
    &\textbf{0.8457}	
    &\textbf{0.8945}	
    &\textbf{681} 
    \\
    \hline
    Improv.
    & 4.8\%
    & 4.4\%
    & -
    & 7.1\%
    & 5.2\%
    & 13.3\%
    \\
    \hline
\end{tabular}
}
}
\end{table*}

\section{experiments}
\subsection{Dataset}
We conducted extensive experiments on two real-world mobility datasets, which are the Geolife and Tencent datasets.
\begin{itemize}[leftmargin=*]
\item \textbf{Geolife}~\cite{zheng2010geolife}: This dataset is compiled by Microsoft Research Asia, which captures the mobility patterns of 182 users from April 2007 to August 2012. Each trajectory represents a sequence of time-stamped GPS points, encompassing latitude, longitude, and altitude data.
\item \textbf{Tencent}: The dataset is collected from Tencent, a prominent social network and location service provider in Beijing, China, during the period from June 1 to June 30, 2018. The trajectory data is derived from GPS locations that are recorded whenever users access location services through the application.
\end{itemize}

\textbf{Pre-processing}: To effectively represent spatial information, the city area is segmented into 10,655 blocks using Beijing’s road infrastructure, with each block representing a unique location averaging approximately 0.265 $km^2$. Consistent with previous research~\cite{chen2019complete}, the temporal resolution adopted is 30 minutes. To maintain dataset reliability, the Tencent dataset excludes trajectories containing less than 34 intervals (around 70\% coverage of a single day). It removes users who lack trajectory records spanning at least five days. For the Geolife dataset, trajectories with fewer than 12 intervals and users who do not have a minimum of five days of mobility data are similarly discarded. Table~\ref{table:datasets} summarizes the processed mobility datasets.

\subsection{Baselines}
We evaluate ten baseline models on two mobility datasets, encompassing both rule-based and data-driven approaches. Linear, History, Top, and RF are rule-based algorithms designed for trajectory recovery, leveraging insights into human mobility patterns.

\begin{itemize}[leftmargin=*]
\item \textbf{Linear}~\cite{hoteit2014estimating}: This approach assumes that missing trajectory points can be estimated using a linear function. As a rule-based method, it relies on a strong underlying assumption about mobility patterns.
\item \textbf{History}~\cite{li2019reconstruction}: This approach straightforwardly leverages historical trajectories by assigning the most frequently visited location at each time slot as the recovered position.
\item \textbf{Top}~\cite{liu2016predicting}: This approach determines the recovery location for each user by selecting the most frequently visited location across the entire training set, making it a frequency-based method.
\item \textbf{RF}~\cite{li2019reconstruction}: The random forest method leverages various trajectory features for recovery. A random forest classifier is trained using attributes such as trajectory entropy, radius, missing time gaps, and the spatial locations before and after these gaps.
\end{itemize}

Data-driven algorithms like LSTM, BiLSTM, DeepMove, Bi-STDDP, AttnMove, and CSDI use machine learning models to capture more complex mobility features:
\begin{itemize}[leftmargin=*]
\item \textbf{LSTM}~\cite{liu2016predicting}: This approach employs mobility prediction for trajectory recovery by leveraging LSTM networks to model sequential transitions, enabling the inference of the next trajectory point.
\item \textbf{BiLSTM}~\cite{zhao2018prediction}: This approach enhances the standard LSTM network by incorporating a bidirectional structure, allowing it to utilize historical and future observations. Leveraging all available locations as spatiotemporal constraints improves trajectory point recovery beyond the limitations of unidirectional modeling.
\item \textbf{DeepMove}~\cite{feng2018deepmove}: This trajectory recovery method is based on mobility prediction, utilizing a multimodal embedding recurrent neural network. By jointly embedding multiple factors influencing human movement, it effectively captures complex sequential transitions.
\item \textbf{Bi-STDDP}~\cite{xi2019modelling}: This approach simultaneously models spatiotemporal dependencies and user preferences. By leveraging the locations visited immediately before and after the target time period, it effectively reconstructs missing trajectory points.
\item \textbf{AttnMove}~\cite{xia2021attnmove}: This method incorporates both intra-trajectory and inter-trajectory attention mechanisms to enhance the modeling of user movement patterns. Additionally, it exploits periodic trends in long-term historical data to improve trajectory recovery.
\item \textbf{PeriodicMove}~\cite{sun2021periodicmove}: This methodology leverages an attention mechanism integrated with graph neural networks to effectively model complex location transitions, hierarchical periodic behaviors, and temporal variations inherent in human movement patterns.
\item \textbf{TrajGDM}~\cite{chu2024simulating}: This approach employs a diffusion-based trajectory modeling framework to learn the underlying mobility patterns within a trajectory dataset.
\end{itemize}

\subsection{Experimental Settings}
To enhance training and evaluation while accounting for unknown missing patterns, 20\% of the observations in both datasets are randomly masked as recovery targets. The data is then partitioned chronologically, with the first 70\% used for training, the next 10\% for validation, and the final 20\% for testing.

For evaluation, we employ three key metrics: \textit{Recall}, \textit{MAP}, and \textit{Distance}.  
\textit{Recall} quantifies the model's effectiveness in recovering ground-truth locations, averaged across all test cases. A score of 1 signifies perfect recovery, while 0 indicates complete failure.  
\textit{MAP} evaluates the overall accuracy of location rankings by assessing how effectively the locations are ordered, with higher values indicating superior performance. \textit{Distance} quantifies the spatial discrepancy between the predicted location's center and the actual location, where smaller values signify better accuracy and enhanced prediction quality.

\subsection{Experiment Results}

\subsubsection{Overall Performance}
To evaluate the reconstructed mobility trajectories, we utilize three primary metrics and present the outcomes in Table~\ref{tab:performance}. Based on these experimental results, we derive the following insights:

Firstly, rule-based methods consistently underperform across all three metrics. While they offer high interpretability by modeling mobility patterns with explicit physical meanings, they struggle to capture intricate, time-dependent, and higher-order mobility features that cannot be strictly defined through predefined rules.  

Secondly, data-driven approaches achieve notable improvements by directly learning mobility patterns from real-world data without relying on rigid assumptions. This enables them to better model complex mobility behaviors. Methods that effectively incorporate historical trajectories or utilize attention mechanisms show particularly strong performance, especially in cases of sparse trajectory data. Since sequential dependencies may be weak in such scenarios, all observed locations should be considered equally. For instance, if a user is recorded at 7:00 AM and 3:00 PM, accurately inferring their 9:00 PM location may require prioritizing the 7:00 AM observation, as it is more likely to correspond to a residential location rather than a commuting destination.  

Thirdly, on the Geolife dataset, our method improves \textit{Recall} and \textit{MAP} by \textbf{6.95\%–7.28\%} over the best baseline. Although its \textit{Distance} metric does not surpass the linear model, this may be due to the dataset’s small size. However, the improvements in the other two metrics confirm its effectiveness in trajectory recovery. On the Tencent dataset, our approach surpasses the best baselines in \textit{Recall} and \textit{MAP} by \textbf{8.44\%–10.61\%}, while also reducing \textit{Distance} by \textbf{27.09\%}. The lower performance gain on Geolife compared to Tencent is likely attributed to the smaller user base, which makes it more challenging to extract periodic patterns and leverage group-level mobility features for individual trajectory recovery.  

Overall, our method outperforms both rule-based and data-driven approaches, demonstrating its effectiveness in integrating individual and group mobility patterns, capturing periodic behaviors from historical data, and utilizing spatiotemporal correlations in current trajectories for enhanced recovery. These advantages highlight its superior performance in trajectory reconstruction.


\begin{figure}[t]
\centering
\subfloat[Geolife]{\includegraphics[width=.23\textwidth]{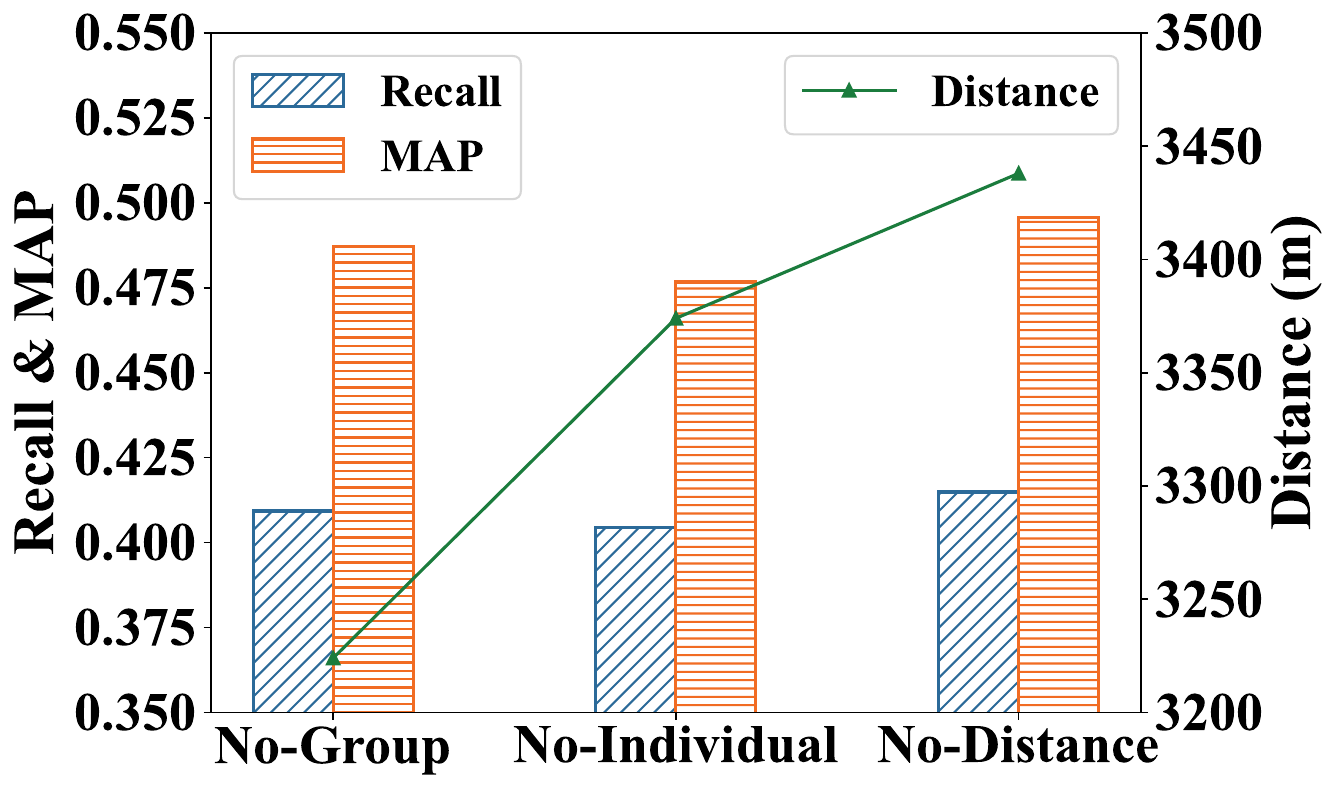}}
\hfil
\subfloat[Tencent]
{\includegraphics[width=.22\textwidth]{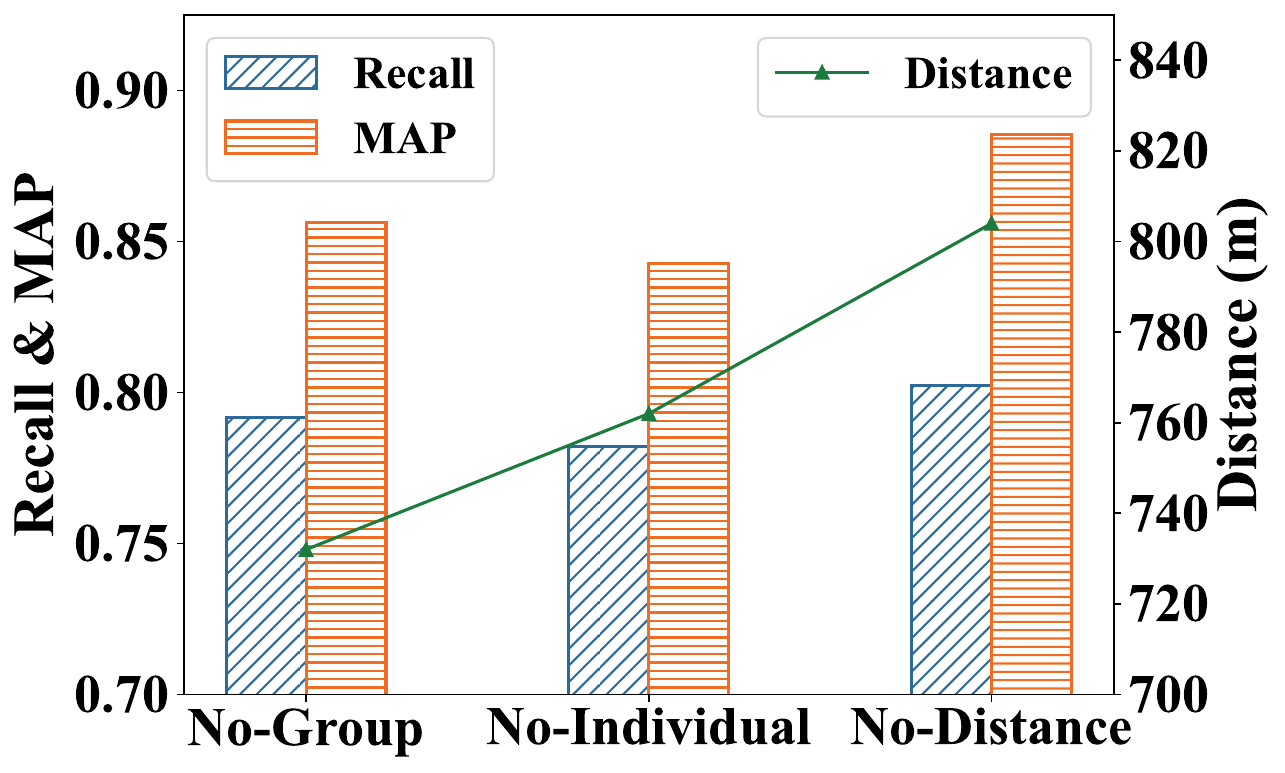}}
\caption{Impact of each component on two datasets.} 
\label{fig:geolife}
\end{figure}

\subsubsection{Ablation Study}
To assess the contribution of each module, we performed ablation experiments by systematically removing them and analyzing the impact on performance, with results shown in Figure~\ref{fig:geolife}. "No-Group" eliminates the group preference extraction module, simplifying the generation of location embeddings. "No-Individual" removes the individual preference extraction module, applying diffusion and denoising directly to fixed location embeddings from the group preference module. "No-Distance Loss" excludes the distance-aware loss $\mathcal{L}_d$ from the objective function.  

Performance degradation confirms that removing the individual preference extraction module has the most significant impact, emphasizing its role in leveraging historical and current data to capture individual mobility preferences. In contrast, omitting the group preference module has a relatively more minor effect, as the location embeddings it generates still enrich the individual preference module, enabling a more comprehensive integration of individual and group mobility patterns.  

Additionally, excluding $\mathcal{L}_d$ most notably affects the \textit{Distance} metric, reinforcing its role in enforcing spatial continuity constraints by penalizing excessive deviations between consecutive trajectory points.

\begin{figure}[t]
\centering
\subfloat[Head and layer]{\includegraphics[width=.23\textwidth]{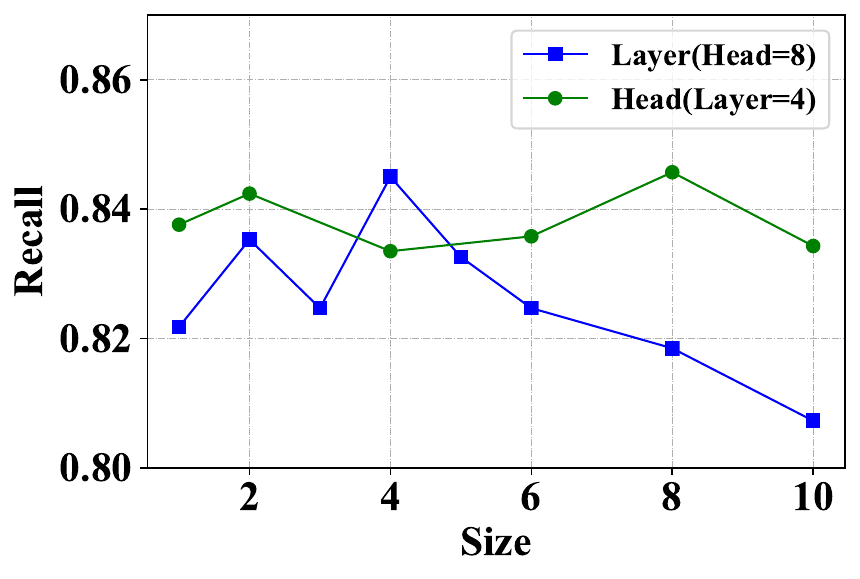}}
\hfil
\subfloat[The weight of distance-aware loss]
{\includegraphics[width=.23\textwidth]{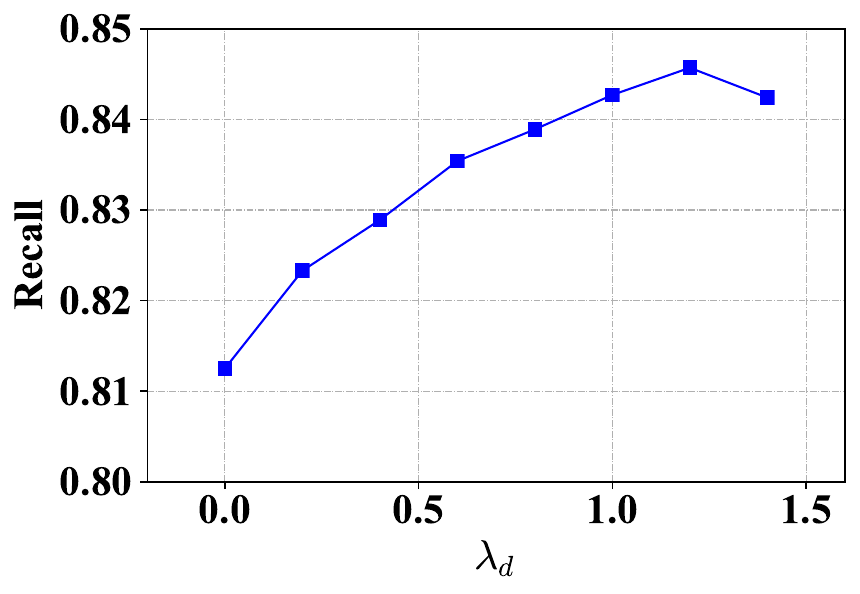}}
\caption{Sensitivity of hyper-parameters.}
\label{fig:hyper-parameters}
\end{figure}

\subsubsection{Sensitivity of Hyper-parameters}
We analyzed the sensitivity of key hyperparameters, including the number of attention heads and layers, and the weight $\lambda_d$ of the distance-aware loss. The results presented are based on the Tencent dataset, with similar trends observed in Geolife.  

Figure~\ref{fig:hyper-parameters}(a) shows the variation in \textit{Recall} when adjusting the number of layers from 1 to 8 while keeping the number of heads constant. Performance peaked at four layers. Similarly, when varying the number of heads from 1 to 10 while maintaining a fixed number of layers, no clear trend emerged, indicating a minor impact of this parameter. As a result, we set the number of layers to 8 and the number of heads to 4.  

For the weight $\lambda_d$ of the distance-aware loss, increasing $\lambda_d$ consistently improved the \textit{Distance} metric, while *Recall* reached its optimal value at $\lambda_d = 1.2$. Therefore, we selected $\lambda_d = 1.2$ for the Tencent dataset.

\subsubsection{Robustness Analysis}
To evaluate the robustness of our model under varying missing ratios, we conducted experiments with results summarized in Table~\ref{table:robustness}. Three key observations emerge from this analysis.  
First, as the missing ratio increases, the performance of both the best baseline, AttnMove, and our model declines across all three metrics, confirming the negative impact of missing data on trajectory recovery.  
Second, across all missing ratios, our model consistently outperforms AttnMove, demonstrating its ability to maintain superior performance under different levels of data sparsity.  
Finally, even with an 80\% missing ratio, our model still surpasses AttnMove when tested with a 20\% missing ratio, underscoring its robustness in handling highly sparse datasets.  

The enhanced robustness of our model is attributed to its ability to jointly capture both group and individual preferences while leveraging a diffusion model to integrate these factors into mobility intention modeling. In contrast, AttnMove primarily relies on individual trajectory information without incorporating these additional design elements. In extreme cases where users have limited trajectory data, our model can still effectively infer mobility intentions by utilizing group-level preferences. By considering a broader spectrum of information, our approach significantly improves performance in scenarios with high data sparsity, reinforcing its robustness.

\begin{table}[t]
\caption{Performance w.r.t missing ratios on Tencent
dataset.}
\label{table:robustness}
\begin{center}
\begin{tabular}{c|c|cccc}

 \hline
 \multicolumn{2}{c|}{Missing Rate}  &20\% &40\% &60\% &80\%\\
 \hline
\multirow{3}*{AttnMove} & Recall & 0.7646 & 0.7502 & 0.7267 & 0.7145\\
         & MAP & 0.8249  & 0.8056 & 0.7731 & 0.7628\\ 
         & Distance & 934  & 1012 & 1146 & 1195\\  \hline
\multirow{3}*{DiffMove} & Recall & 0.8457  & 0.8301 & 0.8076 & 0.7823\\ 
         & MAP & 0.8945  & 0.8612 & 0.8428 & 0.8364\\ 
         & Distance & 681  & 745 & 802 & 835\\ 
\hline
\end{tabular}
\end{center}
\end{table}


\section{Conclusion}
We propose a novel mobility trajectory recovery method named DiffMove, which captures the group tendency and individual preference simultaneously.
To capture the group mobility tendency from the group view, we construct a group tendency graph and fuse the group tendency into location representations with graph embedding.
To model the individual mobility preference from the individual view, 
we first capture individual mobility preferences from their historical and current trajectories and then utilize a diffusion model to refine the spatio-temporal distribution of these trajectories.
We conduct extensive experiments on two real-world datasets, and the results verify the effectiveness of our DiffMove model.
In the future, we plan to extend our framework. Semantic-aware trajectory recovery can be realized by considering the functional properties of the visited locations.

\bibliographystyle{IEEEtran}

\bibliography{sample-base}

\vfill

\end{document}